\documentclass[11pt]{article}

\usepackage[preprint]{acl}

\usepackage{times}
\usepackage{latexsym}

\usepackage[T1]{fontenc}

\usepackage[utf8]{inputenc}

\usepackage{microtype}

\usepackage{inconsolata}

\usepackage{graphicx}

\title{Scalable Prompt Routing via Fine-Grained Latent Task Discovery}

\author{Yunyi Zhang, Soji Adeshina, Sheng Guan, Ashwin Ganesh, Zhen Han,\\
\bf Vassilis N. Ioannidis, Huzefa Rangwala, George Karypis\\
Amazon Web Services\\
\texttt{zhyunyi@amazon.com}}

\usepackage{amsmath, amsfonts, amsthm, xspace, color}
\usepackage{booktabs} 

\usepackage{multirow, tabularx} 
\usepackage{enumitem}
\usepackage{graphicx}
\usepackage{makecell}
\usepackage{caption}
\usepackage{subcaption}
\PassOptionsToPackage{table}{xcolor}

\definecolor{darkgreen}{rgb}{0.0, 0.5, 0.0}

\newcommand{\eg}{e.g.\xspace} 
\newcommand{\nop}[1]{}

\newtheoremstyle{exampstyle}
  {0pt} 
  {0pt} 
  {\itshape} 
  {1em} 
  {\bfseries} 
  {.} 
  {.5em} 
  {} 

\theoremstyle{exampstyle}

\def \L {\mathcal{L}}

\def \C {\mathcal{C}}

\def \M {\mathcal{M}}

\def \R {\mathcal{R}}

\def \T {\mathcal{T}}

\newcommand{\Our}{\mbox{FineRouter}\xspace}

\usepackage{dsfont}

\usepackage[most]{tcolorbox}
\usepackage{soul}
\sethlcolor{pink}

\begin{document}
\maketitle

\begin{abstract}
Prompt routing dynamically selects the most appropriate large language model from a pool of candidates for each query, optimizing performance while managing costs. As model pools scale to include dozens of frontier models with narrow performance gaps, existing approaches face significant challenges: manually defined task taxonomies cannot capture fine-grained capability distinctions, while monolithic routers struggle to differentiate subtle differences across diverse tasks. We propose a two-stage routing architecture that addresses these limitations through automated fine-grained task discovery and task-aware quality estimation. Our first stage employs graph-based clustering to discover latent task types and trains a classifier to assign prompts to discovered tasks. The second stage uses a mixture-of-experts architecture with task-specific prediction heads for specialized quality estimates. At inference, we aggregate predictions from both stages to balance task-level stability with prompt-specific adaptability. Evaluated on 10 benchmarks with 11 frontier models, our method consistently outperforms existing baselines and surpasses the strongest individual model while incurring less than half its cost.
\end{abstract}

\section{Introduction}

Large language models (LLMs) exhibit diverse capabilities across different task types, with no single model consistently outperforming all others. This heterogeneity motivates \textit{prompt routing}, aiming to dynamically select the most appropriate model from a candidate pool for each query to optimize performance while managing computational costs. As model pools expand to include dozens of powerful candidates, a fundamental challenge emerges: how can routing methods accurately distinguish fine-grained capability differences across models and task types at scale?

Existing prompt routing approaches face significant limitations when scaling to large model pools with narrow performance gaps. Most methods either rely on manually defined coarse-grained task taxonomies~\cite{nvidia_llm_router} or train monolithic routers that predict model quality across all prompt types~\cite{ong2025routellm,feng-etal-2025-ipr,chen2024frugalgpt,ding2024hybrid}. The former approach becomes infeasible as manual taxonomy design cannot keep pace with the nuanced strengths of LLMs, while the latter struggles to capture fine-grained distinctions when a single estimator must differentiate subtle capability differences across diverse tasks. For instance, within the broad category of ``mathematics,'' models may exhibit vastly different performance on symbolic algebraic manipulation versus contextual word problems, yet coarse categorization treats these uniformly. Furthermore, when routing among frontier models with narrow performance gaps, the routing task becomes substantially more challenging, requiring the system to identify subtle task-model affinity patterns that determine which model is \textit{best} for a given prompt.

We propose a two-stage routing architecture, \Our, that leverages automated fine-grained task discovery and task-aware quality estimation. Rather than forcing a monolithic model to handle all distinctions simultaneously, we explicitly infer latent task structure, allowing specialized components to focus on specific task types. Our Stage 1 develops an offline graph-based clustering method that automatically discovers fine-grained task types from training data. For each discovered task type, we adaptively select top candidate models and train a classifier to efficiently assign task types to input prompts during inference. Stage 2 employs a mixture-of-experts quality estimation architecture where task-specific prediction heads are invoked based on the assigned task type. This design enables specialized routing knowledge for each task type while maintaining computational efficiency. At inference, our router combines complementary signals from both stages, balancing task-level stability with instance-level adaptability.

We evaluate our approach on 10 diverse benchmarks spanning various tasks, routing among 11 state-of-the-art frontier models including Claude-Sonnet-4.5, DeepSeek-R1, Llama-4-Maverick, and Qwen3-235B. Our method consistently outperforms existing routing baselines and achieves superior performance compared to any individual model, including surpassing the strongest candidate while incurring less than half its inference cost. Ablation studies confirm that both stages contribute meaningfully to overall performance, with fine-grained task discovery providing more effective routing signals than coarse-grained taxonomies. Case studies reveal that our clustering method successfully identifies meaningful task distinctions that align with known model capabilities while also discovering unexpected niche domains.

Our main contributions are as follows:
\begin{itemize}[leftmargin=*,noitemsep,topsep=0pt]
    \item A scalable automated task discovery method that combines semantic and performance-based signals to identify fine-grained task types from large-scale training data.
    \item A task-aware routing architecture employing mixture-of-experts quality estimation with specialized prediction heads that leverage discovered task structure for more accurate model selection.
    \item Comprehensive evaluation on 10 benchmarks with 11 frontier models as candidates, demonstrating consistent improvements over baselines and superior cost-performance tradeoffs compared to single-model deployment.
\end{itemize}
\section{Preliminaries}

\textbf{Prompt Routing} Let $\M = \{M_1, M_2, \ldots, M_n\}$ denote a set of $n$ candidate large language models. Given an input prompt $p$, the goal of prompt routing is to select the most appropriate model $M^* \in \M$ that optimizes a desired objective.

\smallskip
\noindent\textbf{Quality-Based Routing} Most existing prompt routing methods frame this as a single-class classification task which predicts one LLM that performs the best on the input prompt~\cite{ong2025routellm,feng2025graphrouter}. However, we argue that as the LLMs become more powerful, there are increasingly more cases where multiple models perform similarly well. These subtle distinctions in their performance cannot be captured by a coarse classification output and thus makes the classification setting brittle. Therefore, we adopt Quality-Based Routing~\cite{feng-etal-2025-ipr} which turns routing into a regression task. Formally, assume there is a quality function $Q^*: (p, r) \rightarrow \mathbb{R}$ that assigns a quality score to each pair of prompt and LLM-generated response, $r = M(p)$. Empirically $Q^*$ can be any task-specific evaluation functions or general reward models~\cite{liu2025skywork}. Then the best-performing model selection can be defined as:
\vspace{-0.5em}
\begin{equation}\label{eq:gt-best}
\small
    M^*(p) = \arg\max_{M_i \in \M} Q^*(p, M_i(p)).
\vspace{-0.5em}
\end{equation}
However, evaluating all models at inference time is computationally prohibitive and counterproductive to the routing formulation. Therefore, the routing problem requires learning a quality estimator $\tilde{Q}$ that predicts the best model without generating responses from all candidates,
\vspace{-0.5em}
\begin{equation}
\small
    \R(p) = \arg\max_{M_i \in \M} \tilde{Q}(p, M_i).
\vspace{-0.5em}
\end{equation}

\noindent\textbf{Task-Aware Routing Objective} As $n$ grows large, learning an accurate quality estimator $\tilde{Q}$ becomes challenging. Different models excel at different task types, and a monolithic estimator struggles to capture these fine-grained distinctions. This motivates our approach of discovering latent task structure to enable task-aware quality estimation. To address this challenge, we introduce the concept of latent task types. Let $\T = \{t_1, t_2, \ldots\}$ represent a set of discovered task types. Our goal is to learn both a task assignment function $t: p \rightarrow \mathcal{T} \cup \{\emptyset\}$ and a task-aware routing function,
\vspace{-0.5em}
\begin{equation}
\small
    \R_t(p) = \arg\max_{M_i \in \M} \tilde{Q}_t(p, M_i).
\vspace{-0.5em}
\end{equation}
that leverages task-specific knowledge to improve routing accuracy while maintaining computational efficiency at inference time.

\begin{figure*}[t]
    \centering
    \vspace*{-0.5em}
    \centerline{\includegraphics[width=0.97\textwidth]{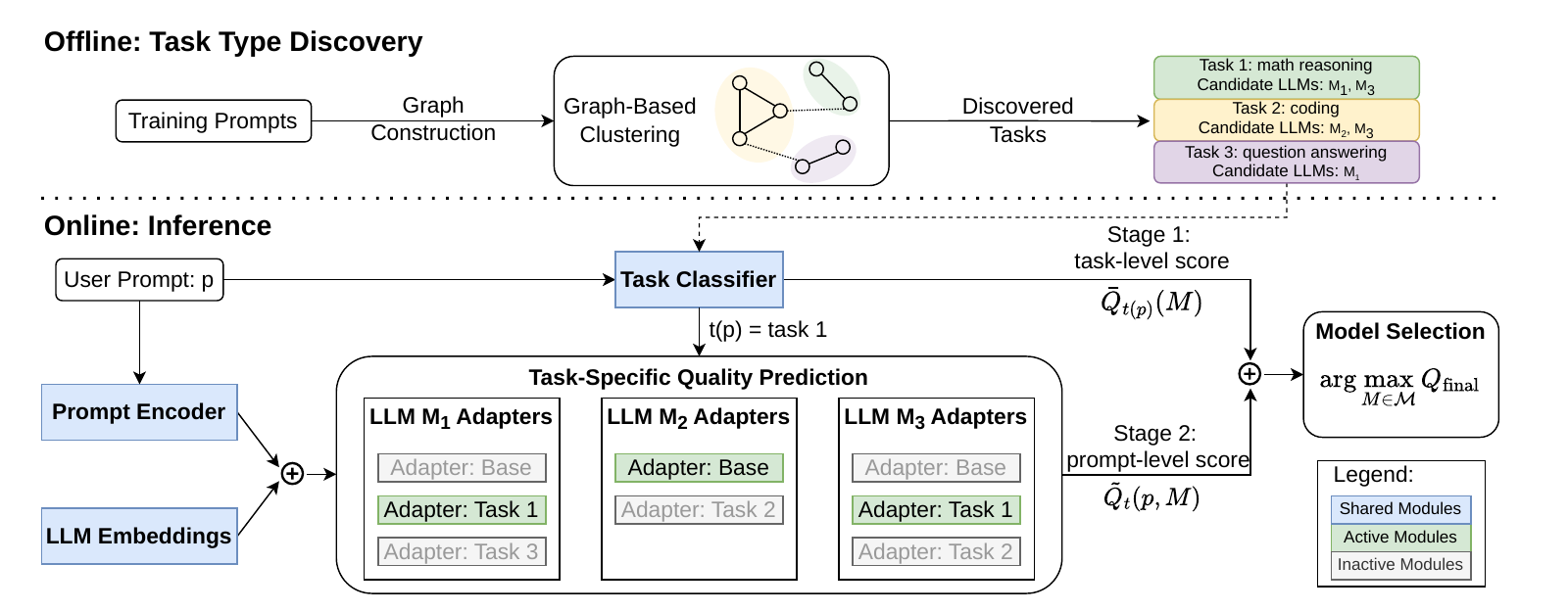}}
    \caption{Overview of \Our. \textbf{Top}: Offline task type discovery via graph-based clustering, producing fine-grained tasks with candidate LLMs per task. \textbf{Bottom}: Online inference where the task classifier assigns prompts to discovered tasks, enabling task-specific adapter activation in the MoE router. Final model selection aggregates task-level scores (Stage 1) with prompt-specific quality predictions (Stage 2).}
    \label{fig:framework}
    \vspace*{-1em}
\end{figure*}

\section{Methodology}

Figure~\ref{fig:framework} shows an overview of our two-stage routing architecture. Stage 1 identifies fine-grained task types through graph-based clustering and trains a classifier to assign new prompts to discovered tasks. Stage 2 employs an MoE quality estimation model with task-specific prediction heads. At inference, predictions from both stages are aggregated to produce the final model selection. 

\subsection{Stage 1: Task Type Discovery and Matching}

The first stage of our routing architecture automatically discovers fine-grained task types from training data and learns to match new prompts to these discovered tasks. Unlike existing works that rely on pre-defined coarse-grained task taxonomies~\cite{nvidia_llm_router, feng2025graphrouter}, our method automates the discovery of fine-grained task structure directly from data. This automation is crucial for two reasons: (1) manual task taxonomy design becomes infeasible as model pools scale to hundreds of candidates with nuanced strengths, and (2) data-driven discovery can reveal latent task distinctions that may not be apparent through manual categorization but are critical for effective model differentiation.

\subsubsection{Task Type Discovery}\label{sec:task-type}

Given a training set covering diverse prompts from different sources, we first use an LLM to generate a concise sentence describing the task for each prompt.\footnote{We show our prompt in Appendix~\ref{app:prompt}.} These task descriptions provide semantic representations that capture the nature of each prompt. We then apply a graph-based clustering method that combines two complementary signals: (1) semantic similarity between task descriptions, and (2) similarity in model preference patterns as reflected by ranked lists of preferred LLMs.

For a prompt $p$, let $\pi_p = [M_1, M_2, \ldots, M_n]$ denote a ranked list of LLMs from the model pool $\M$, where models are ordered by decreasing preference on their responses according to the quality function, $Q^*(p, M_i(p))$. The rank of model $M$ in list $\pi_p$ is denoted as $\text{rank}_{\pi_p}(M)$.

\smallskip
\noindent\textbf{Graph Construction} We construct a sparse prompt graph where nodes represent individual prompts and edges encode both semantic and performance-based similarity. First, we identify $k$-nearest neighbors for each prompt based on cosine similarity between task description embeddings. For each candidate edge, we compute a pairwise Rank Biased Overlap (RBO)~\cite{rbo} score between the ranked lists of preferred LLMs for the two prompts. We filter out edges with RBO scores below a threshold $\tau$, retaining only pairs that exhibit similar model preferences. For the remaining edges, we set edge weights to the geometric mean of normalized cosine similarity and RBO scores after min-max normalization.

\smallskip
\noindent\textbf{Iterative Clustering} We apply Leiden community detection~\cite{traag2019louvain} to identify prompt clusters. For each detected community $C = \{p_1, p_2, \ldots, p_{|C_j|}\}$, we (1) compute a cluster center through median pooling of its task description embeddings and (2) construct a combined ranked list of preferred LLMs using rank fusion across all prompts in the cluster using the mean reciprocal rank (MRR) score,
\vspace{-0.5em}
\begin{equation}
\small
    \text{MRR}_{C}(M) = \frac{1}{|C|}\sum_{p_i \in C} \frac{1}{\text{rank}_{\pi_p}(M) + \epsilon},
\vspace{-0.5em}
\end{equation}
where $\epsilon$ is a constant (typically $\epsilon=60$) that reduces the impact of high-rank differences. This summarization enables recursive application of the community detection algorithm. 

This iterative process continues for $\ell$ iterations to refine the clusters. We only keep the clusters from the final iterations, where each cluster $\C_i$ represents a discovered task type $t_i$ consisting of semantically similar prompts that share similar preferred LLMs. Importantly, our method only clusters prompts that form meaningful task communities. Prompts that do not fit into any cluster are treated as not belonging to a specific task type.

\smallskip
\noindent\textbf{Candidate Model Selection} For each discovered task cluster $\C_i$, we identify a small set of top candidate LLMs $\mathcal{L}_i$ that are most likely to perform well on that specific task. We use the same rank fusion process to combine the ranked lists of preferred LLMs across all prompts within the cluster. Instead of using a fixed top-$k$ parameter, we adaptively choose the number of candidate LLMs to maximize the coverage of the candidate set over the cluster's preferred models.
Specifically, let $\text{Cov}(\L_i)$ be the frequency of prompts within the cluster whose most preferred LLM appears in $\L_i$.
Then we incrementally increase the number of candidate models until $\text{Cov}(\L_i)$ exceeds a predefined threshold $\delta$. This adaptive selection ensures that each task type is associated with a focused set of strong candidate models.

\subsubsection{Task Type Classifier}

To match prompts with discovered task types at inference time, we train a text classifier that predicts matching scores between prompts and task types, denoted as $t: p \rightarrow \mathcal{T} \cup \{\emptyset\}$. The classifier employs a bi-linear matching architecture consisting of two components: (1) a prompt encoder initialized from a pre-trained text encoder, and (2) task type encodings initialized with embeddings of LLM-generated summarizing task descriptions for each prompt cluster. This architecture enables efficient computation of matching scores between prompt and task type representations. 

Given the large number of target classes (typically hundreds of discovered tasks), we fine-tune the classifier in a multi-label setting using binary cross-entropy loss. This formulation allows prompts to potentially match multiple task types with varying confidence scores, providing flexibility in task assignment during inference.

\subsection{Stage 2: Task-Aware Dynamic Router}

With the task type classifier from Stage 1, we can now assign any prompt to one of the discovered fine-grained task types (or to no specific task if all matching scores are low). Building on these task assignments, Stage 2 employs a task-aware routing mechanism. Instead of training a single monolithic router across all prompt types, we leverage the discovered task structure to enable specialized routing decisions through a mixture-of-experts architecture. Specifically, we utilize a mixture-of-experts quality estimation architecture where specialized prediction heads are invoked based on the predicted task type of the incoming prompt.

\subsubsection{Model Architecture}

Our router model consists of three main components: (1) a prompt encoder initialized from a pre-trained transformer-based encoder, (2) an LLM embedding layer that maps model IDs to model-specific representations, and (3) a Quality Estimation (QE) layer with task-aware prediction heads. Input prompts and LLMs are encoded with (1) and (2) respectively, which are then concatenated and passed to the QE layer to predict an estimated quality score $\tilde{Q}_t(p, M)$.

The QE layer implements a mixture-of-experts architecture with two types of MLP-based prediction heads. First, we maintain $|\M|$ general adapters (MLPs), with each adapter predicting the quality score for one specific model. These general adapters learn to estimate each model's expected response quality based on global knowledge acquired from the entire training data, providing baseline predictions applicable across all prompt types.

Second, for each discovered task type $t_i$ with the selected candidate models $\L_i$ identified in Stage 1, we initialize $|\L_i|$ task-specific quality prediction adapters. Because these candidate LLMs are most likely to perform well on the corresponding task, the task-specific adapters can learn specialized knowledge that better predicts their performance on this particular task type. All adapter heads share the same prompt encoder and LLM embeddings, ensuring efficient parameter usage while enabling specialized predictions. 

If a prompt is assigned to task type $t_i$, we invoke the task-specific adapters for models in $\L_i$ and the general adapters for all other models in $\M - \L_i$. This hybrid invocation strategy combines the benefits of task-specific expertise with comprehensive coverage: the task-specific adapters provide refined predictions for the most promising candidates based on learned task patterns, while the general adapters ensure that potentially strong models outside the selected candidates are still considered, preventing the router from prematurely excluding viable options.

\subsubsection{Model Training}

We train the router model to mirror its inference-time behavior. First, we re-label the entire training set using the task type classifier obtained from Stage 1. During training, task-specific prediction heads are trained only on prompts assigned to their corresponding task type, allowing each expert to specialize in its designated task domain. We optimize all prediction heads using mean squared error (MSE) loss between the predicted quality scores $\tilde{Q}_t(p, M)$ and the ground-truth quality scores $Q^*(p, M(p))$.

For more effective training, we adopt a two-phase approach. We first train the base model consisting of the prompt encoder, LLM embedding layer, and general quality prediction heads on all training data. Then, we fine-tune the task-specific prediction heads while freezing the prompt encoder and LLM embeddings on the task-type-labeled training data. This staged training strategy ensures that the shared representations remain stable while task-specific experts learn to refine predictions for their specialized domains.

\subsection{Inference}

At inference time, we deploy both stages to produce the final routing decision through a two-step process that combines task-based prior knowledge with prompt-specific quality estimation.

First, we apply the task type classifier to assign a task type $t_i \in \T \cup \{\emptyset\}$ to the incoming prompt $p$. If the prompt is assigned to a specific task $t_i$, we invoke the corresponding task-specific adapters for models in $\L_i$ along with general adapters for models in $\M - \L_i$ to obtain quality estimates $\tilde{Q}_t(p, M)$ for all models. If no task assignment is made (i.e., $t(p) = \emptyset$), we use only the general adapters to predict quality scores across all models.

To leverage complementary information from both stages, we aggregate their predictions through a weighted combination. Stage 1 provides task-based prior knowledge through the aggregated quality scores of each model on the assigned prompt cluster, denoted as $\bar{Q}_{t_i}(M)$, which represents the median quality score of model $M$ across all prompts in cluster $\C_i$. Note that for prompts where $t(p) = \emptyset$, we construct an 'Others' cluster from all such training prompts and compute their aggregated median scores $\bar{Q}_{\emptyset}(M)$ as stage-1 scores. Stage 2 provides fine-grained, prompt-specific quality estimates $\tilde{Q}_t(p, M)$. We normalize both sets of scores to the range $[0, 1]$ using min-max normalization (denoted as $\text{norm}(\cdot)$) and compute the final routing score as:
\vspace{-0.5em}
\begin{align} 
\label{eq:final-score}
\small
\begin{split}
Q_{\text{final}}(p, M) &= \alpha \cdot \text{norm}(\tilde{Q}_t(p, M))\\
                        &+ (1-\alpha) \cdot \text{norm}(\bar{Q}_{t(p)}(M)), 
\vspace{-0.7em}
\end{split}
\end{align} 
where $\alpha \in [0, 1]$ controls the relative weight between prompt-specific and task-based predictions. The final model selection is then: 
\vspace{-0.5em}
\begin{equation} 
\small
R_t(p) = \arg\max_{M \in \mathcal{M}} Q_{\text{final}}(p, M).
\vspace{-0.7em} 
\end{equation}
This aggregation strategy enables the router to benefit from both the stability of task-level patterns and the adaptability of prompt-specific predictions, resulting in more robust routing decisions.

Importantly, this inference process maintains computational efficiency: the task type classifier requires only a single forward pass, and the adaptive activation of prediction heads ensures that the effective model size during inference remains constant regardless of the number of discovered tasks.
\section{Experiments}

\begin{table*}[!t]
\centering
\caption{Performance comparison across 10 benchmarks. Best model and best router scores are \textbf{boldfaced}.}
\label{table:main-res}
\vspace*{-0.5em}
\scalebox{0.75}{
\begin{tabular}{l|cccccccccc|cc}
\toprule
\multirow{2}{*}{\bf Models}    & \multicolumn{10}{c}{\bf Per-Task Evaluation} & \multicolumn{2}{|c}{\bf Overall} \\
\cmidrule{2-11} \cmidrule{12-13}
& NQ & Triv-QA & Com-QA & mmlu & arc & OB-QA & gsm8k & math & hum-eval & mbpp & Ave & Qual \\
\midrule
\multicolumn{13}{l}{\textit{Candidate Models}}\\
Llama-3.3-70B & 57.6 & 23.2 & 76.0 & 87.4 & 90.5 & 90.7 & 89.2 & 94.0 & 21.1 & 40.0 & 67.0 & 0.446 \\
Llama-4-Maverick & 53.8 & 22.7 & 79.9 & 93.0 & \bf 98.8 & 95.7 & \bf 94.0 & \bf 95.2 & 21.1 & 86.7 & 74.1 & 0.580\\
Claude-Haiku-4.5 & 46.6 & 21.6 & 79.8 & 85.2 & \bf 98.8 & 95.5 & 83.8 & 88.1 & \bf 73.7 & 86.7 & 76.0 & 0.572\\
Claude-Sonnet-4.5 & 59.4 & 24.7 & \bf 85.7 & \bf 94.7 & 97.6 & \bf 97.4 & 88.7 & 91.1 & 63.2 & \bf 93.3 & \bf 79.6 & \bf 0.621\\
Mistral-Small & 50.3 & 22.3 & 59.2 & 74.1 & 69.1 & 67.2 & 74.1 & 25.1 & 31.6 & 26.7 & 50.0 & 0.409\\
Mistral-Large & 56.0 & 24.8 & 75.8 & 70.5 & 92.9 & 85.1 & 84.4 & 71.0 & 57.9 & 23.3 & 64.2 & 0.489\\
DeepSeek-v3 & \bf 60.7 & 25.7 & 70.7 & 89.2 & 91.7 & 84.0 & 91.9 & 89.0 & 36.8 & 73.3 & 71.3 & 0.412\\
DeepSeek-R1 & 60.3 & \bf 26.0 & 81.0 & 92.0 & \bf 98.8 & 94.4 & 90.4 & 91.6 & 47.4 & 86.7 & 76.9 & 0.551\\
Qwen3-32B & 39.5 & 18.7 & 78.9 & 85.3 & 96.4 & 90.3 & 86.6 & 75.5 & 21.1 & 33.3 & 62.6 & 0.528\\
Qwen3-235B-A22B & 55.1 & 23.7 & 85.2 & 93.9 & 91.7 & 90.9 & 90.6 & 93.3 & 52.6 & 83.3 & 76.0 & 0.530\\
GPT-OSS-120B & 48.3 & 21.6 & 78.7 & 89.6 & 97.6 & 93.5 & 79.1 & 88.4 & 26.3 & 80.0 & 70.3 & 0.362\\
\midrule
\multicolumn{13}{l}{\textit{Routers}}\\
kNN & 56.2 & 24.4 & 79.8 & 91.9 & 97.6 & 94.6 & 90.3 & 92.2 & 42.1 & 66.7 & 73.6 & 0.620\\
MLP & 53.6 & 23.8 & 60.2 & 74.7 & 70.2 & 68.2 & 75.4 & 59.5 & 63.2 & 80.0 & 62.9 & 0.449\\
RouteLLM & 55.0 & 22.1 & 77.1 & 89.2 & 97.6 & 92.9 & 82.6 & 87.5 & 26.3 & 43.3 & 67.4 & 0.387\\
RouterDC & 50.3 & 22.4 & 76.4 & 89.0 & 91.7 & 90.5 & 82.9 & 88.0 & 26.3 & 46.7 & 66.4 & 0.353\\
GraphRouter & 48.3 & 21.7 & 78.7 & 90.1 & \bf 98.8 & 93.5 & 79.7 & 89.6 & 26.3 & 80.0 & 70.7 & 0.386\\
IPR & 57.6 & 24.3 & 81.7 & 92.2 & 97.7 & 95.3 & 90.4 & 93.6 & 47.4 & 83.3 & 76.3 & 0.646\\
\Our & \bf59.0 & \bf 24.9 & \bf 83.3 & \bf 93.4 & \bf 98.8 & \bf 95.7 & \bf 91.6 & \bf 93.8 & \bf 68.4 & \bf 90.0 & \bf 79.9 & \bf 0.652\\
\bottomrule
\end{tabular}
}
\vspace*{-1em}
\end{table*}

\subsection{Experimental Setup}

\noindent\textbf{Datasets}
We evaluate our approach on 10 benchmark datasets that cover a wide range of natural language understanding and reasoning tasks: question answering (NQ~\cite{kwiatkowski-etal-2019-natural}, TriviaQA~\cite{joshi-etal-2017-triviaqa}, CommensenseQA~\cite{talmor-etal-2019-commonsenseqa}), multiple-choice (MMLU~\cite{hendryckstest2021, hendrycks2021ethics}, ARC-Challenge~\cite{allenai:arc}, OpenBookQA~\cite{mihaylov-etal-2018-suit}), mathematical reasoning (GSM8K~\cite{cobbe2021gsm8k}, MATH~\cite{hendrycksmath2021}), and code generation (HumanEval~\cite{chen2021evaluating}, MBPP~\cite{austin2021program}). We split the combined dataset into 278,977 training samples, 34,872 development samples, and 34,873 test samples.

\smallskip
\noindent\textbf{Model Candidates}
We evaluate our routing approach across a diverse set of 11 recent state-of-the-art language models spanning multiple model families and capability profiles. Our candidate pool includes models from the Llama family (\textsc{Llama-3.3-70B}~\cite{meta2024llama33}, \textsc{Llama-4-Maverick}\footnote{\url{https://ai.meta.com/blog/llama-4-multimodal-intelligence/}}), Anthropic's Claude series (\textsc{Claude-Haiku-4.5}\footnote{\url{https://www.anthropic.com/claude/haiku}}, \textsc{Claude-Sonnet-4.5}\footnote{\url{https://www.anthropic.com/claude/sonnet}}), Mistral AI models\footnote{\url{https://docs.mistral.ai/getting-started/models}} (\textsc{Mistral-Large}, \textsc{Mistral-Small}), DeepSeek (\textsc{DeepSeek-v3}~\cite{deepseek2024v3}, \textsc{DeepSeek-R1}~\cite{deepseek2025r1}), Qwen3 models~\cite{qwen2025qwen3} (\textsc{Qwen3-32B}, \textsc{Qwen3-235B-A22B-Thinking}), and OpenAI's open-source model \textsc{GPT-OSS-120B}~\cite{openai2025gptoss}. 

Unlike previous works that primarily focus on smaller open-source models, we intentionally select frontier models to mimic real-world deployment scenarios where users seek to optimally leverage all available models. Routing among such high-performing models presents a significantly more challenging task, as the performance gaps between models are narrower and more nuanced.

\smallskip
\noindent\textbf{Baselines} 
We compare against several routing methods: 
(1) kNN: selects models based on embedding similarity to training examples, 
(2) MLP: classifies query embedding to the most proper candidate LLM, 
(3) RouteLLM~\cite{ong2025routellm}: matrix factorization based routing that learns latent factors for prompts and models, 
(4) RouterDC~\cite{chen2024RouterDC}: dual contrastive learning based routing, 
(5) GraphRouter~\cite{feng2025graphrouter}: graph-based routing using prompt relationships to tasks and LLMs,
(6) IPR~\cite{feng-etal-2025-ipr}: a quality estimation-based routing approach by fine-tuning prompt and LLM encoding with per-LLM quality prediction adapters.

For all compared methods, we use a state-of-the-art reward model \textsc{Skywork-Reward-V2-Llama-3.1-8B}~\cite{liu2025skywork} as the reference quality function $Q^*(p, M(p))$ after normalizing to range $[0, 1]$. Therefore, for classification-based methods, the ground truth label is just $M^*(p)$ for each training prompt (Eq.~\ref{eq:gt-best}). While existing works typically use the downstream evaluation metric as quality function (\eg, Exact Match for QA tasks), most of such metrics are binary and induce lots of ties in scores for stronger candidate models. Selecting the ``best performing model'' for each training prompt will be biased to ``first-appearing candidate,'' leading to all classification-based routers overfit to select one single model for all test samples. For complete implementation details, see Appendix~\ref{app:imp-detail}.

\smallskip
\noindent\textbf{Evaluation Metrics} We report the task-specific evaluation metric for each dataset following~\cite{llmrouter2025} with their macro average, and a quality score using $Q^*(\cdot)$.

\begin{table*}[!t]
\centering
\caption{Ablation study comparing stage-1 only, stage-2 only, and full two-stage architecture.}
\label{table:ablation}
\vspace*{-0.5em}
\scalebox{0.7}{
\begin{tabular}{l|cccccccccc|cc}
\toprule
\multirow{2}{*}{\bf Models}    & \multicolumn{10}{c}{\bf Per-Task Evaluation} & \multicolumn{2}{|c}{\bf Overall}                   \\
\cmidrule{2-11} \cmidrule{12-13}
& NQ & Triv-QA & Com-QA & mmlu & arc & OB-QA & gsm8k & math & hum-eval & mbpp & Ave & Qual \\
\midrule
Stage 1 Only\\
\quad Embedding & 58.5 & 24.8 & \bf 84.5 & 92.5 & 95.2 & 95.5 & 88.7 & \bf 94.3 & 42.1 & 86.7 & 76.3 & 0.609\\
\quad Coarse CLS & \bf 60.3 & \bf 25.9 & 80.4 & 92.5 & \bf 98.8 & 95.0 & \bf 92.8 & 93.0 & \underline{63.2} & \bf 93.3 & 79.5 & 0.623\\
\quad Fine CLS & \underline{59.3} & 25.1 & 82.7 & \underline{93.3} & \bf 98.8 & \bf 95.7 & \underline{91.8} & 93.6 & \underline{63.2} & \bf 93.3 & 79.7 & 0.637\\
Stage 2 Only & 57.4 & 24.3 & 81.8 & 92.2 & 97.6 & 95.5 & 90.7 & \underline{94.0} & 52.6 & 83.3 & 77.0 & 0.647\\
\Our & 59.0 & \underline{24.9} & \underline{83.3} & \bf 93.4 & \bf 98.8 & \bf 95.7 & 91.6 & 93.8 & \bf 68.4 & 90.0 & \bf 79.9 & \bf 0.652\\
\bottomrule
\end{tabular}
}
\vspace*{-0.75em}
\end{table*}

\begin{figure*}[t]
    \centering
    \begin{minipage}{0.32\textwidth}
        \centering
        \includegraphics[width=\textwidth]{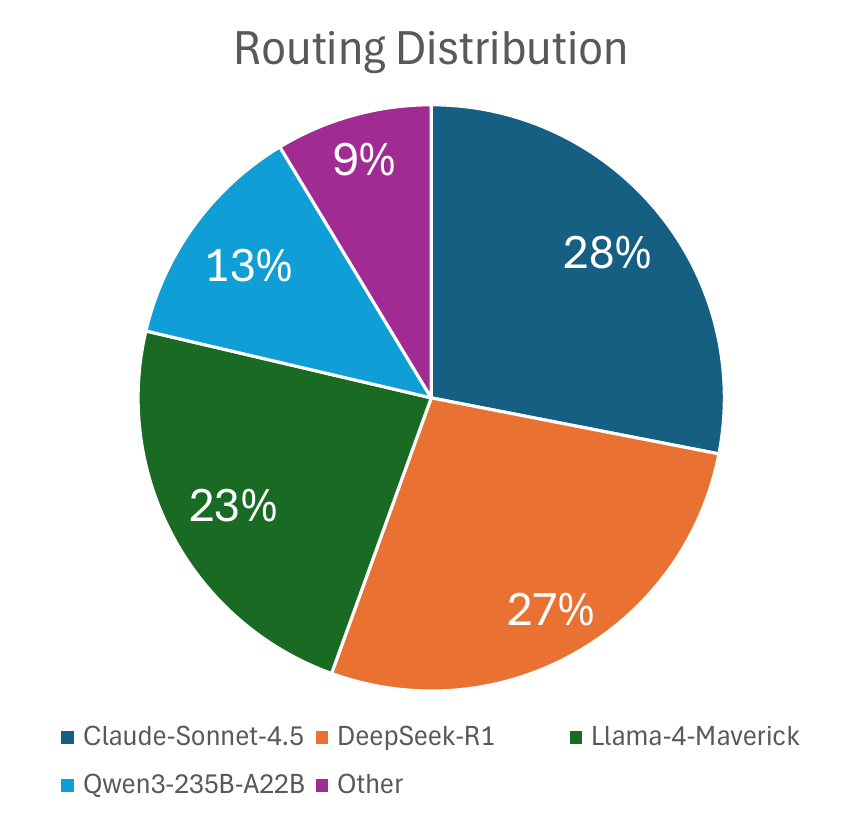}
    \end{minipage}
    \hfill
    \begin{minipage}{0.58\textwidth}
        \centering
        \includegraphics[width=\textwidth]{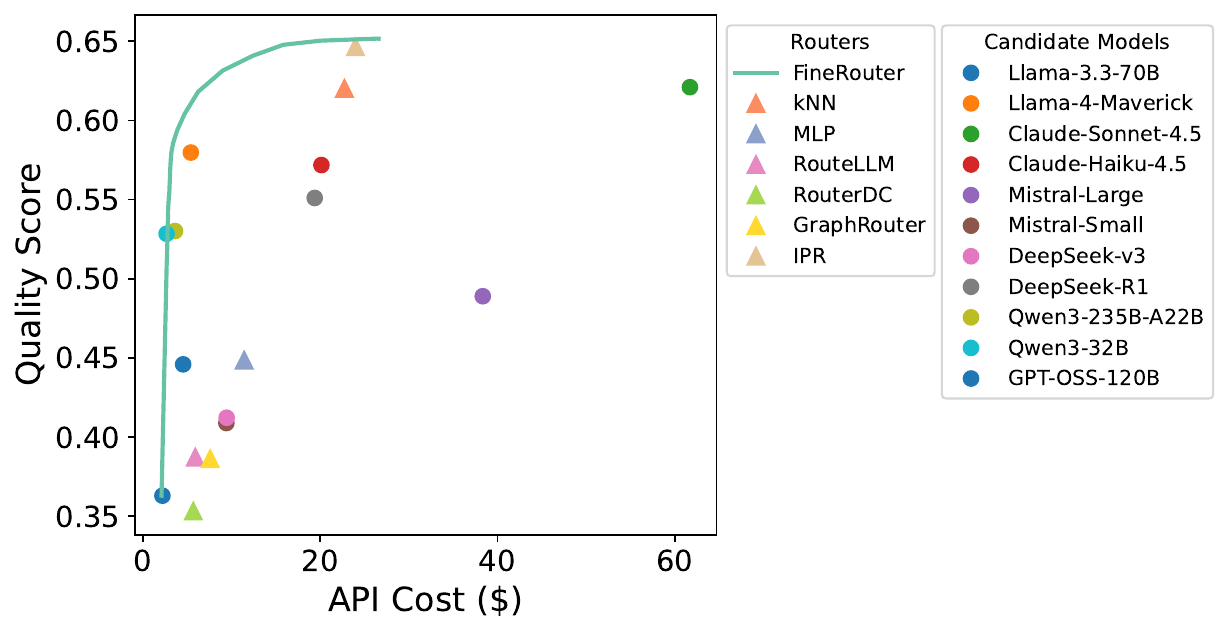}
    \end{minipage}
    \vspace*{-1em}
    \caption{(a) Routing distribution of FineRouter across 11 candidate models. (b) Cost-performance comparison. FineRouter (curve) outperforms baseline routers (triangles) and individual LLMs (circles).}
    \label{fig:analysis}
    \vspace*{-1.25em}
\end{figure*}

\subsection{Experiment Results}

Table~\ref{table:main-res} presents the performance of \Our compared to baselines and individual candidate models across 10 diverse benchmarks. First, no single LLM consistently achieves the best performance across all tasks, which strongly motivates the value of prompt routing. Second, baseline routing methods struggle to outperform strong individual models, likely because our candidate pool consists of state-of-the-art frontier models with narrow performance gaps, making it substantially harder to distinguish subtle capability differences compared to prior work focusing on smaller models with larger performance disparities. Third, \Our achieves the strongest overall performance, consistently outperforming baselines across the majority of tasks. This demonstrates that explicit fine-grained task discovery and task-aware routing successfully leverage the complementary strengths of different frontier models.

\subsection{Routing Behavior Analysis}

Figure~\ref{fig:analysis}(a) illustrates the routing distribution of \Our across the 11 candidate models on the test set. The router demonstrates balanced utilization across multiple high-performing models, with Claude-Sonnet-4.5 receiving 28\%, DeepSeek-R1 27\%, Llama-4-Maverick 23\%, and Qwen3-235B 13\% of prompts. This reflects the router's ability to identify and leverage each model's strengths across different task types, rather than over-relying on a single model.

Figure~\ref{fig:analysis}(b) presents the cost-performance tradeoff comparing \Our, baseline routers, and individual LLMs. We plot the average quality score based on ground truth $Q^*$ against the estimated total API cost\footnote{As of Feb 2026 through AWS Bedrock API: \url{https://aws.amazon.com/bedrock/pricing/}} on all test samples. To obtain the cost-performance curve for routers, we follow~\citep{feng-etal-2025-ipr} to vary a tolerance parameter from 0 to 1, which controls the minimum performance the selected model should achieve comparing to the best score, all based on $Q_{\text{final}}(p, M)$ (Eq.~\ref{eq:final-score}). Our two-stage router consistently dominates baseline routers across all cost points while achieving superior performance compared to individual LLMs. Notably, \Our achieves better performance than the strongest candidate Claude-Sonnet-4.5 at less than half its cost, validating that task-aware routing enables better resource allocation.
We also include a task classifier analysis in Appendix~\ref{app:classifier}.

\begin{table*}[t]
\centering
\caption{Representative discovered task types showing fine-grained distinctions and model-specific strengths.}
\label{tab:case-studies}
\vspace*{-0.5em}
\small
\begin{tabular}{>{\centering}m{1.5cm}m{8.5cm}m{3cm}}
\toprule
\textbf{Class ID} & \textbf{Task Description} & \textbf{Selected Candidates}\\
\midrule
109 & Answer multiple-choice questions requiring inference, detail identification, numerical calculation, and statement verification. & \makecell[l]{DeepSeek-v3} \\
\midrule
267 & Formal symbolic mathematics: Solve algebra, geometry, number theory, linear algebra, and calculus problems. & \vspace{-1em}\makecell[l]{Llama-4-Maverick, \\ Llama-3.3-70B}\\
\midrule
161 & \vspace{0.5em}Telephone area code queries: geographic locations, regional identifiers, and telecom history. & \vspace{-1em}\makecell[l]{Qwen3-235B, \\ GPT-OSS-120B, \\ DeepSeek-R1}\\
\bottomrule
\end{tabular}
\vspace*{-1em}
\end{table*}

\subsection{Ablation Studies}

To understand the contribution of each component, we conduct ablation studies comparing several variants. We include 3 variants for stage-1 only ablations: 
\textbf{Embeddings} uses prompt embeddings to match test prompts to the cluster centers and routes it to the top candidate model identified for the corresponding task.
\textbf{Coarse CLS} uses a pretrained classifier with 10 pre-defined coarse-grained task types~\cite{nvidia_llm_router}, combined with our rank fusion method on our training data to determine recommended models for each task type. 
\textbf{Fine CLS} refers to our stage-1 task type classifier and routes each prompt to the top candidate LLM for the predicted task.
\textbf{Stage-2 Only} uses the predicted scores from the stage-2 dynamic router without merging it with stage-1 scores.

Table~\ref{table:ablation} presents the ablation results. First, comparing Coarse CLS with Fine CLS demonstrates that our automatically discovered fine-grained task types provide more effective routing signals than manually pre-defined coarse-grained categories. Second, Stage-1-only approaches achieve competitive performance by leveraging discovered task structure to filter candidate models, validating the effectiveness of our task discovery method. Third, the full two-stage architecture consistently outperforms both Stage-1-only and Stage-2-only variants, demonstrating that task-aware specialized prediction heads provide complementary fine-grained routing signals beyond task-based candidate filtering alone. These results confirm that both stages contribute meaningfully to the overall routing performance, with explicit task modeling enabling more accurate model selection decisions.

\subsection{Case Studies: Task Discovery}

Our Stage 1 clustering discovers 332 fine-grained task types with an average of 3.55 candidates per task, reducing the model pool to $\sim$32\%.
Table~\ref{tab:case-studies} presents representative examples.
\textbf{Task 109} represents reading comprehension requiring numerical reasoning, with DeepSeek-v3 as the sole recommended model, aligned with its known strength in multi-step reasoning.
\textbf{Task 267} reveals fine-grained distinctions within math: our method picks Llama-4-Maverick and Llama-3.3-70B for symbolic mathematics, while general word problems benefit from Claude-Sonnet-4.5 and DeepSeek-v3. This distinction would be missed by coarse-grained categorization that simply treats both as "math."
\textbf{Task 161} exemplifies unexpected niche domains our method discovers: telephone area code queries combining geographic and telecom knowledge. Such fine-grained discoveries enable our router to make more informed decisions rather than treating all factual questions uniformly.
\section{Related Work}

Prompt routing aims to dynamically select the most appropriate model from a candidate pool for each query, balancing performance and cost. 
Early work by \citet{chen2024frugalgpt} introduced FrugalGPT, which cascades multiple LLMs and uses a learned scoring function to determine when to stop the cascade. 
HybridLLM~\cite{ding2024hybrid} employs a BERT-based encoder to optimize cost-quality trade-offs by routing "easy" queries to smaller models and "hard" queries to larger models. 
RouteLLM~\cite{ong2025routellm} trains router models using human preference data and matrix factorization or BERT-based classifier to learn latent factors for prompts and models. 
Zooter~\cite{lu-etal-2024-routing} uses reward model scores as supervision signals and trains routers with RankNet loss, incorporating tag-based label enhancement to reduce reward model noise. 
RouterDC~\cite{chen2024RouterDC} employs dual contrastive learning to train query-based routers that capture fine-grained distinctions between LLM capabilities. 
GraphRouter~\cite{feng2025graphrouter} constructs a heterogeneous graph connecting queries to pre-defined coarse-grained task categories and LLMs, using edge prediction for routing. However, it relies on manually specified task taxonomies that cannot scale to capture the nuanced capability differences among frontier models.
\citet{feng-etal-2025-ipr} developed IPR, a quality-constrained routing framework with modular architecture using per-LLM quality prediction adapters and user-controlled tolerance parameters for explicit cost-quality trade-offs. 
Our work differs from all prior approaches in two key ways: (1) we automatically discover task structure rather than relying on pre-defined categories, and (2) we use discovered tasks to specialize prediction heads rather than just filtering
candidates.
\section{Conclusion}

We present a two-stage routing architecture, \Our, that addresses the challenge of scaling prompt routing to large pools of frontier models with narrow performance gaps. Our key contribution is the automated discovery of fine-grained latent task structure through graph-based clustering and a mixture-of-experts router that provides specialized quality estimates for each task type.
Evaluated on 10 diverse benchmarks with 11 frontier models, our method consistently outperforms existing routing baselines and surpasses the strongest individual model while incurring less than half its inference cost.
Our work demonstrates that explicit modeling of latent task structure enables more accurate and scalable routing decisions, particularly as model pools expand to include dozens of powerful candidates with subtle capability differences.

\section*{Limitations}

While our approach demonstrates strong performance, several limitations merit discussion. First, our task discovery relies on LLM-generated task descriptions and reward model scores as supervision signals. The quality of discovered task types depends on the capability of the LLM used for description generation (we use Claude-Sonnet-4.5) and the accuracy of the reward model (Skywork-Reward-V2). These dependencies may introduce biases or miss task distinctions that are not well-captured by current reward models, particularly for specialized domains or creative tasks where reward modeling remains challenging.

Second, our graph-based clustering requires several hyperparameters (k-nearest neighbors, RBO threshold $\tau$, coverage threshold $\delta$, number of iterations $\ell$) that were tuned on our training data. While we provide default values that work well across our benchmarks, optimal settings may vary for different domains or model pools. The clustering process also assumes that prompts with similar semantic descriptions and model preferences form meaningful task communities, which may not hold for all prompt distributions.

Third, our two-stage architecture requires training both a task classifier and a mixture-of-experts router, which involves computational overhead during the training phase. The method also requires access to responses from all candidate models on training data to compute quality scores, which may not be feasible in all deployment scenarios.

Fourth, our approach currently handles text-only prompts. It is a promising direction for future work to extend task discovery to multimodal inputs where task structure may emerge from visual or audio features.

\bibliography{custom}

\appendix

\begin{figure*}[t!]
\centering
\begin{tcolorbox}[
  enhanced,
  title=Task Description Generation,
  attach boxed title to top center={yshift=-3mm,yshifttext=-1mm},
  colback=gray!15,
  colframe=gray!75!black,
  colbacktitle=gray!40,
  coltitle=black,
  fonttitle=\bfseries,
  boxed title style={size=small,colframe=gray!50!black},
  boxrule=0.5pt,
  arc=7pt,
  outer arc=7pt,
  left=10pt,
  right=10pt,
  top=10pt,
  bottom=10pt
]
Your task is to generate a short description of the task that a user's prompt intends to do.

You are allowed to perform chain-of-thought or thinking but the final answers should be in <task> tags with the following instructions:

- The description should be one short sentence.

- The description should cover different aspects of the tasks, including but not limited to task types (e.g., classification, generation, translation, ...), topic of interests (e.g., technology, health, finance, ...), domain of knowledge.

- Do not respond to the provided prompt. Your task is to generate a description of what it tries to do

- Do not include explanations, reasoning, context, or commentary of any kind in the task description

- Do not preface or conclude your answer with statements like "Based on my knowledge..." or "The task is..." in the task description

- Try to always output an answer

- Format your response exactly as follows:

\bigskip
<task>
task description
</task>
\bigskip

Now, please generate the task description for the following prompt: \hl{\{$p$\}}

\end{tcolorbox}
\caption{Prompt given to the LLM to generate task descriptions for offline task type discovery (Sect.~\ref{sec:task-type}).}
\label{fig:prompt-task-desc}
\end{figure*}

\newpage
\section{Prompt for Task Description Generation}
\label{app:prompt}

Figure~\ref{fig:prompt-task-desc} shows the prompt we used to generate domain and task descriptions for training prompts during the offline task type discovery process in Section~\ref{sec:task-type}.

\section{Implementation Details} 
\label{app:imp-detail}
For Stage 1 task discovery, we use \textsc{Claude-Sonnet-4.5} to generate task descriptions for each training prompt. We construct the prompt graph with $k=5$ nearest neighbors based on embeddings from Sentence Transformer \textsc{all-MiniLM-L6-v2}~\cite{reimers-gurevych-2019-sentence}. We set the RBO threshold $\tau=0.4$ and apply Leiden community detection for $\ell=3$ iterations. The coverage threshold $\delta$ for adaptive candidate selection is set to 0.8.

The task type classifier is initialized from \textsc{stella\_en\_400M\_v5}~\cite{zhang2025jasperstelladistillationsota} and fine-tuned for 10 epochs with a learning rate of 2e-5 and per-device batch size of 8.
For Stage 2, the prompt encoder is initialized from \textsc{Qwen3-Embedding-0.6B}~\cite{qwen3embedding}. The LLM embedding layer has dimension 512. Each quality estimation adapter consists of a 2-layer MLP with hidden dimensions 512. We first train the base model for 10 epochs, then fine-tune task-specific adapters for another 10 epochs while freezing other parts. All models use per-device batch size 8.
All model training is done on 8 NVIDIA A100 GPUs.
Training the full pipeline takes approximately 17 hours on 8 NVIDIA A100 GPUs (around 135 GPU-hours): 6 hours for the base model and 11 hours for task-specific adapter fine-tuning.

At inference, we set the aggregation weight $\alpha=0.5$ in Eq.~\ref{eq:final-score}. We use \textsc{Skywork-Reward-V2-Llama-3.1-8B}~\cite{liu2025skywork} as the reference quality function $Q^*(\cdot)$ for both training and evaluation.


\section{Task Classifier Analysis}
\label{app:classifier}

The task classifier achieves 0.643 macro F1 on a held-out validation set of cluster assignments, which is a 332-class classification problem. On the test set, 71\% of prompts (22,485 out of 31,774) are assigned to a discovered task type. Prompts with task 
assignments achieve a quality score of 0.665, compared to 0.619 for prompts in the "Others" category, confirming that task-specific routing provides meaningful improvements over the general fallback.

\end{document}